\documentclass[letterpaper, 10 pt, journal]{IEEEtran}

\IEEEoverridecommandlockouts

\usepackage{etextools}
\usepackage{amssymb}
\usepackage{float}
\usepackage{makeidx}
\usepackage{amsmath}
\usepackage{epsfig}
\usepackage{epsf}
\usepackage{psfrag}
\usepackage{verbatim}
\usepackage{color}
\usepackage{multirow}
\usepackage{tabularx}
\usepackage{booktabs}
\usepackage[tight,footnotesize]{subfigure}
\usepackage{array}
\usepackage{soul}
\usepackage{footnote}
\usepackage{cite}
\usepackage{dblfloatfix}

\usepackage{graphicx}
\graphicspath{{Figures}}
\DeclareGraphicsExtensions{.pdf,.png}
\usepackage{bigstrut}

\newtheorem{definition}{Definition}

\usepackage{algorithm, setspace}
\usepackage{algpseudocode}
\usepackage{url}
\usepackage{multirow}

\title{{Enabling Robots to Infer how End-Users Teach and Learn through Human-Robot Interaction}}

\author{Dylan P. Losey, \textit{Student Member, IEEE}, and Marcia K. O'Malley, \textit{Senior Member, IEEE} 
\thanks{This work was funded in part by the NSF GRFP-1450681. \newline \indent
	The authors are with the Mechatronics and Haptic Interfaces Laboratory, Department of Mechanical Engineering, Rice University, Houston, TX 77005.
	{(e-mail: dylan.p.losey@gmail.com)}}}

\begin{document}
\maketitle

\begin{abstract}

During human-robot interaction (HRI), we want the robot to understand us, and we want to intuitively understand the robot. In order to communicate with and understand the robot, we can leverage interactions, where the human and robot observe each other's behavior. However, it is not always clear how the human and robot should interpret these actions: a given interaction might mean several different things. Within today's state-of-the-art, the robot assigns a \emph{single} interaction strategy to the human, and learns from or teaches the human according to this fixed strategy. Instead, we here recognize that different users interact in different ways, and so one size does not fit all. Therefore, we argue that the robot should maintain a \emph{distribution} over the possible human interaction strategies, and then {infer how each individual end-user interacts during the task}. We formally define learning and teaching when the robot is uncertain about the human's interaction strategy, and derive solutions to both problems using Bayesian inference. In examples and a benchmark simulation, we show that our personalized approach outperforms standard methods that maintain a fixed interaction strategy.

\end{abstract}


\section{Introduction} \label{sec:intro}

Human-robot interaction (HRI) provides an opportunity for the human and robot to exchange information. The robot can learn from the human by observing their behavior \cite{argall2009}, or teach the human through its own actions \cite{zhu2015}. In applications such as autonomous cars, personal robots, and collaborative assembly, fluent human-robot communication is often necessary.

In order to learn from and teach with interactions, however, the human and robot must correctly interpret the meaning of each other's behavior. Consider an autonomous car following behind a human driven car. If the human car slows down, what should the robotic car infer: is the human teaching the robot to also slow down, or signaling that the robot should pass? When learning from an end-user, the robot needs a model of that end-user's \emph{teaching strategy}, i.e., how the human's actions relate to the information that human wants to convey. Conversely, when teaching the end-user, the robot must model that end-user's \emph{learning strategy}, i.e., how the human interprets the robot's actions. Together, we define the end-user's teaching and learning strategies as their \emph{interaction strategy}.

In the state-of-the-art, the robot assigns a pre-programmed, fixed interaction strategy to every human; each individual end-user is assumed to teach or learn in the same way. Instead:
\begin{center}
\emph{We here recognize that different users have different interaction strategies, and we should infer the current end-user's interaction strategy based on their actions.}
\end{center}
Rather than a single fixed estimate of the human's interaction strategy, we argue that the robot should maintain a distribution (i.e., belief) over the possible human interaction strategies, and update this belief during the task. By reasoning over this belief, the robot can adapt to everyday end-users, instead of requiring each human to comply with its single pre-defined strategy.

{Overall, we make the following contributions:}

\smallskip
\noindent {\textbf{Learning and Teaching with Strategy Uncertainty.} We introduce and formulate two novel problems in human-robot interaction, where the robot must optimally communicate with the human, but the robot is unsure about how the current end-user teaches or learns.}

\smallskip
\noindent {\textbf{Solution with Bayesian Inference.} We derive methods for the robot to learn and teach under strategy uncertainty. We show that---when the robot does not know the end-user's interaction strategy---optimal solutions infer and update a belief over that interaction strategy, resulting in personalized interactions.}

\smallskip
\noindent {\textbf{Simulated Comparison to Current Methods.} Using didactic examples and an inverse reinforcement learning simulation, we compare our proposed approach to robots that reason over a fixed interaction strategy. We also consider practical challenges such as noisy and unmodeled interaction strategies.}

\section{Related Work} \label{sec:related}

\subsection{Robots Learning from Humans}

When a human expert is using interactions to teach a robot, the robot can leverage learning from demonstration (LfD) to understand how it should behave \cite{argall2009}. Most similar to our setting is inverse reinforcement learning (IRL), an instance of LfD where the robot learns the correct reward function from human demonstrations \cite{ng2000,abbeel2004}. Prior works on IRL generally assume that every human has a single, fixed teaching strategy \cite{osa2018}: the human teaches by providing optimal demonstrations, and any sub-optimal human behavior is interpreted as noise \cite{choi2011,losey2018,ramachandran2007,ziebart2008}. Alternatively, robots can also learn about the human while learning from that human. In Nikolaidis \textit{et al}. \cite{nikolaidis2017}, for instance, the robot learns about the end-user's adaptability in addition to their reward function. Building on these works, we will infer the end-user's teaching strategy, so that the robot can more accurately learn from human interactions.

\subsection{Robots Teaching Humans}

Machine teaching---also known as algorithmic teaching---identifies the best way for an expert robot to teach the novice human \cite{zhu2015}. In order to teach optimally, however, the robot must know how the human learns. Recent machine teaching works \cite{liu2017,melo2018,rafferty2016} have addressed this problem by using human feedback to resolve mismatches between the assumed human learning strategy and the user's actual learning strategy. Most related to our research is work by Huang \textit{et al}. \cite{huang2017}, which compares the performance of different models of human learning. These authors generated the optimal teaching examples for each proposed learning strategy, and then used human feedback to identify the single best teaching strategy across all users. Like Huang \textit{et al.} \cite{huang2017}, we here reason over multiple learning strategies, but now we want to infer each user's specific learning strategy based on their individual responses.

\section{Problem Statement} \label{sec:problem}

Consider a human who is interacting with a robot. In this setting, both the human and the robot are agents. Let us assume that one of these agents has a target model, $\theta^* \in \Theta$, which they want to teach to the other agent. Here $\Theta$ is the space of possible target models, and $\theta^*$ is particular behavior that the teacher wants to convey to the learner. For example, the teacher may want to show the learner a better way to complete the current task, or communicate how it will interact in the future. We are interested in how the robot should behave when it is (a) \emph{learning $\theta^*$ from} or (b) \emph{teaching $\theta^*$ to} a human agent.

\subsection{Notation}

Let us denote the robot state as $x$. The human takes action $u$, and the robot takes action $a$; these actions and the state $x$ are observed by both the robot and the human. We use a superscript $t$ to denote the current timestep, so that $x^t$ is the state at time $t$, and $x^{0:t}$ is the sequence of states from the start of the task to the current time $t$. In the context of supervised learning, we can think of $x$ as the input features, and $u$ and $a$ as the output labels assigned by the human and robot, respectively \cite{goodfellow2016}. Here $\Theta$ is a hypothesis space, and $\theta^*$ defines the correct mapping from features to labels.

\subsection{Learning from the Human} \label{sec:problearn}

When the human is the expert---i.e., the human knows $\theta^*$, but the robot does not---the robot should learn from the human. The human wants to teach the robot $\theta^*$, and has a \emph{teaching strategy} $\phi^*$, which determines what actions the human selects to convey $\theta^*$ to the robot. More formally, a teaching strategy $\phi \in \Phi$ relates the setting $(x,\theta)$ to the human action $u$:
\begin{equation} \label{eq:P1}
	\pi(u ~ | ~ x, \theta, \phi)
\end{equation}
Here $\pi \in [0,1]$ is the probability that the human will take action $u$ given $x$, $\theta$, and $\phi$. We point out that (\ref{eq:P1}) is also the human's policy when teaching the robot, and that this policy is parameterized by $\phi$. In other words, if the robot knows the teaching strategy $\phi^*$, then it can leverage (\ref{eq:P1}) to correctly interpret the meaning behind the human's actions.

In practice, however, the robot \emph{does not know} what teaching strategy an end-user will employ. Hence, we argue that the robot should maintain a probability distribution over $\phi$ as it learns from the human. We refer to this problem of learning from the end-user when uncertain about their teaching strategy as \emph{learning with strategy uncertainty}:

\smallskip
\begin{definition}
{(Learning with Strategy Uncertainty). Given a discrete or continuous set of possible teaching strategies $\Phi$ and target models $\Theta$, infer an optimal estimate of $\theta^*$ based on the history of states $x^{0:t}$ and human actions $u^{0:t}$.}
\end{definition}

\subsection{Teaching the Human}

Next we consider the opposite situation, where the robot is the expert, and is trying to teach $\theta^*$ to the human. Here the human agent has some \emph{learning strategy} $\psi^*$, which determines how the human interprets the robot's actions $a$. A learning strategy $\psi \in \Psi$ expresses the relationship (from the human's perspective) between the setting $(x,\theta)$ and the robot action $a$:
\begin{equation} \label{eq:P2}
	\pi(a ~ | ~ x, \theta, \psi)
\end{equation}
In the above, $\pi$ is the human's model of the robot's policy---not necessarily the robot's actual policy---and this model is parameterized by $\psi$. So now, if the robot knows the user's true learning strategy $\psi^*$, the robot can leverage (\ref{eq:P2}) to anticipate how its actions will alter the human's understanding of $\theta^*$.

But, when teaching an actual end-user, the robot \emph{does not know} what learning strategy that specific user has. Similar to before, we therefore argue that the robot should maintain a distribution over the learning strategies $\psi$ when teaching the human. We refer to this problem, where the robot is teaching a user but is unsure about that end-user's learning strategy, as \emph{teaching with strategy uncertainty}:

\smallskip
\begin{definition}
(Teaching with Strategy Uncertainty). Given a {discrete or continuous} set of possible learning strategies $\Psi$ and target models $\Theta$, select the robot action $a^t$ that optimally teaches $\theta^*$ based on the history of states $x^{0:t}$, robot actions $a^{0:t-1}$, and human actions $u^{0:t}$.
\end{definition}

\subsection{Assumptions}

Throughout this work, we will assume that the interaction strategies $\phi^*$ and $\psi^*$ for each individual user are constant, and are not affected by the robot's behavior. Put another way, the robot cannot influence the human's interaction strategy by selecting different actions. {This assumption is consistent with prior HRI research \cite{zhu2015, osa2018}: however, we can also extend our proposed approach to address cases where the human's interaction strategy \emph{does change} by incorporating a forgetting factor or transition model within the Bayesian inference.}

\section{Robot Learning with Strategy Uncertainty} \label{sec:learn}

Within this section we focus on learning from the human, where the robot does not initially know the human's teaching strategy $\phi^*$. Learning here is challenging, because the robot is uncertain about how to interpret the human's actions. First, we demonstrate how the robot can learn from multiple models of the human's teaching strategy. Second, we enable the robot to update its joint distribution over $\phi$ and $\theta$, and simultaneously learn both the human's teaching strategy and target model. We provide an example which compares learning this joint distribution to learning with a single fixed estimate of $\phi^*$.

\subsection{Multiple Teaching Strategies}

The robot starts with a prior $b^0(\theta)$ over what $\theta^*$ is, and updates that belief at every timestep $t$ based on the observed states and actions. The robot's belief over target models is:
\begin{equation} \label{eq:L1}
	b^{t+1}(\theta) = P(\theta ~ | ~ u^{0:t},x^{0:t})
\end{equation}
In other words, $b^{t+1}(\theta)$ is the probability that $\theta = \theta^*$ given the history of observed states and human actions up to timestep $t$. Applying Bayes' rule, and recalling from (\ref{eq:P1}) that the human's actions $u$ are conditionally independent, the robot's Bayesian belief update becomes \cite{russell2009}:
\begin{equation} \label{eq:L2}
	b^{t+1}(\theta) = \frac{b^t({\theta}) \cdot P(u^t~|~x^t;\theta)}{\int_{\Theta} b^t(\xi)\cdot P(u^t~|~x^t;\xi)~d\xi}
\end{equation}
We here used a semicolon to separate the observed variables from the hidden variables. The denominator---which integrates over all possible target models---is a normalizing constant. Omitting this constant, we can more succinctly write (\ref{eq:L2}) as:
\begin{equation} \label{eq:L3}
	b^{t+1}(\theta) \propto b^t({\theta})\cdot P(u^t~|~x^t;\theta)
\end{equation}
where $P(u~|~x;\theta)$ is the robot's \emph{observation model}, i.e., the likelihood that the human takes action $u$ given $x$ and $\theta$.

To correctly learn from the end-user, the robot needs an accurate observation model. We saw in Section~\ref{sec:problearn} that the most accurate observation model is the user's policy $\pi$, which is parameterized by the true teaching strategy $\phi^*$. Within the state-of-the-art, the robot often assumes that the user's policy is parameterized by $\phi^0$, where $\phi^0$ is some estimate of $\phi^*\,$:
\begin{equation} \label{eq:L4}
	P(u^t~|~x^t;\theta) = \pi(u^t~|~x^t;\theta,\phi^0 \,)
\end{equation}
Rather than a \emph{constant point estimate} of the human's teaching strategy, we argue that the robot should maintain a \emph{belief over multiple teaching strategies}. In the simplest case, the robot has a prior $b^0(\phi)$ over what $\phi^*$ is, but does not update this belief between timesteps. Here the observation model becomes:
\begin{equation} \label{eq:L5}
	P(u^t~|~x^t;\theta) = \int_{\Phi} \pi(u^t~|~x^t;\theta,\phi) \cdot b^0(\phi)~d\phi
\end{equation}
Note that (\ref{eq:L4}) is a special case of (\ref{eq:L5}) where $b^0(\phi^0\,) = 1$. When learning with (\ref{eq:L5}), the robot does not interpret human actions in the context of just one teaching strategy. Instead, the robot considers what the action $u$ implies for each possible teaching strategy, and then learns across these strategies. {We can think of (\ref{eq:L5}) as the best fixed learning strategy when $b^0$ is known.}

\subsection{Inferring a Joint Belief} \label{sec:joint}

Now that we have introduced learning with multiple teaching strategies, we can solve learning with strategy uncertainty (Definition~1). Here we not only want to learn the target model $\theta^*$, but we also recognize that the robot is uncertain about $\phi^*$. Let us define the robot's joint belief $b(\theta,\phi)$ over the target models $\theta \in \Theta$ and teaching strategies $\phi \in \Phi$ to be:  
\begin{equation} \label{eq:L6}
	b^{t+1}(\theta,\phi) = P(\theta,\phi~|~u^{0:t},x^{0:t})
\end{equation}
Again leveraging Bayes' rule and conditional independence:
\begin{equation} \label{eq:L7}
	b^{t+1}(\theta,\phi) \propto b^t(\theta,\phi) \cdot P(u^t~|~x^t;\theta,\phi)
\end{equation}
where $P(u ~ | ~ x; \theta, \phi)$ is the conditional probability of human action $u$ given $x$, $\theta$, and $\phi$. But this is the same as (\ref{eq:P1}), so that:
\begin{equation} \label{eq:L8}
	b^{t+1}(\theta,\phi) \propto b^t(\theta,\phi) \cdot \pi(u^t~|~x^t;\theta,\phi)
\end{equation}
Using (\ref{eq:L8}), we learn about \emph{both the human's target model and the human's teaching strategy} from $x^{0:t}$ and $u^{0:t}$. 

Let us compare the observation model for this joint learning rule to the observation models from (\ref{eq:L4}) and (\ref{eq:L5}). If we rewrite (\ref{eq:L8}) into the form of (\ref{eq:L3}), we obtain the observation model:
\begin{equation} \label{eq:L9}
	P(u^t ~ | ~ x^t;\theta) = \int_{\Phi} \pi(u^t ~ | ~ x^t; \theta, \phi) \cdot b^t(\phi ~|~ \theta) ~d\phi
\end{equation}
where the belief over teaching strategies given $\theta$ is:
\begin{equation} \label{eq:L10}
	b^{t}(\phi ~ | ~ \theta) = P(\phi ~ | ~ u^{0:t-1},x^{0:t-1};\theta)
\end{equation}
Intuitively, a robot implementing (\ref{eq:L9}) and (\ref{eq:L10}) reasons across multiple teaching strategies when learning from the end-user, and also updates its belief over these teaching strategies every timestep. We find that the observation model (\ref{eq:L5}) is a special case of (\ref{eq:L9}) when the robot never updates ${b^0(\phi ~ | ~ \theta) = b^0(\phi)}$, i.e., if the robot's belief over teaching strategies is constant. Accordingly, (\ref{eq:L4}) is a special case of (\ref{eq:L9}) by extension. {Our analysis shows that inferring a joint belief over $\theta$ and $\phi$ both generalizes prior work and is an optimal learning rule.}


\subsection{Learning Example} \label{sec:learnex}

\begin{figure}[t]

\vspace{0.5em}

	\begin{center}
		\includegraphics[width=0.7\columnwidth]{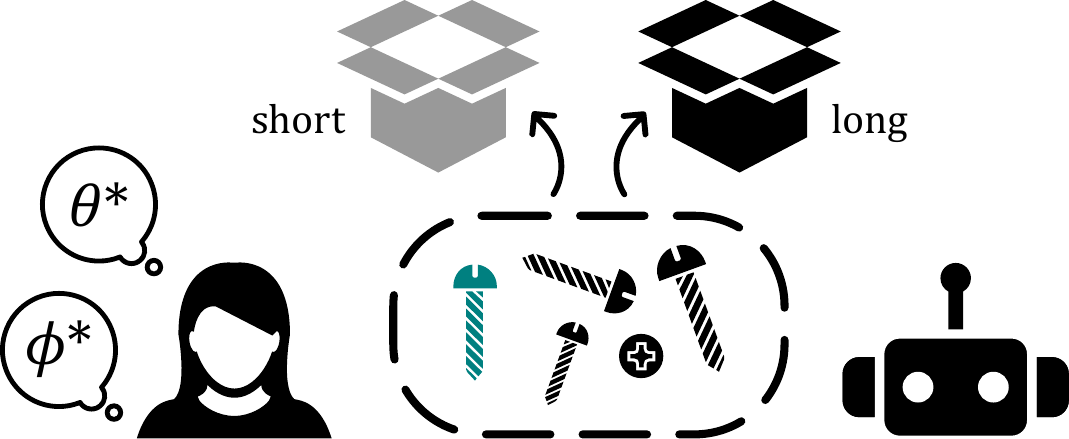}

		\caption{A human and robot are interacting to sort screws of different lengths. The human teacher indicates one screw that they consider short (highlighted), and then the robot learner sorts the screws into the short and long boxes. The robot does not know which screws are short  \emph{a priori}, and also does not know the end-user's teaching strategy. For instance, the human could be indicating a short screw at random, or purposely selecting the longest of the short screws.}

		\label{fig:example1}
	\end{center}

\vspace{-2em}

\end{figure}

To demonstrate how the proposed observation models affect the robot's learning, we here provide an example simulation. Consider the sorting task in Fig.~\ref{fig:example1}, where the robot is attempting to learn the right threshold classifier from the human. At each timestep $t$, the human action $u$ indicates one screw that should be classified as short; the robot then classifies the remaining screws without additional guidance. Let $\theta^*$ be the correct decision boundary, and let the robot's reward equal the total number of screws classified correctly. We can think of this example as an instance of inverse reinforcement learning \cite{abbeel2004}, where the robot learns the true objective $\theta^*$. 

Importantly, we include two different teaching strategies $\phi \in \Phi$ that the human might use. Within the first strategy, $\phi_1$, the human noisily indicates the short screw closest to $\theta^*$, so that ${\pi(\phi_1) \propto \exp\{-\frac{1}{2} \cdot | \theta^* - u|\}}$. Within the second strategy, $\phi_2$, the user indicates a short screw uniformly at random, so that $\pi(\phi_2) \propto 0.9$ if $u\leq\theta^*$ or $\pi(\phi_2) \propto 0.1$ otherwise. Each end-user leverages one of these two teaching strategies; however, the robot does not know which.

\smallskip
\noindent \textbf{Observation Models}. We compare (\ref{eq:L4}), (\ref{eq:L5}), and (\ref{eq:L9}). Let $\phi_1$ denote a robot that learns with (\ref{eq:L4}), and assumes $\phi_1 = \phi^*$ for all users. Similarly, $\phi_2$ is a robot that assumes $\phi_2 = \phi^*$. \emph{Prior} denotes a robot with observation model (\ref{eq:L5}), and \emph{Joint} leverages our proposed approach (\ref{eq:L9}). Finally, $\phi^*$ is an ideal robot that knows the teaching strategy for each individual user.

\begin{figure}[t]

\vspace{0.7em}

	\begin{center}
		\includegraphics[width=1.0\columnwidth]{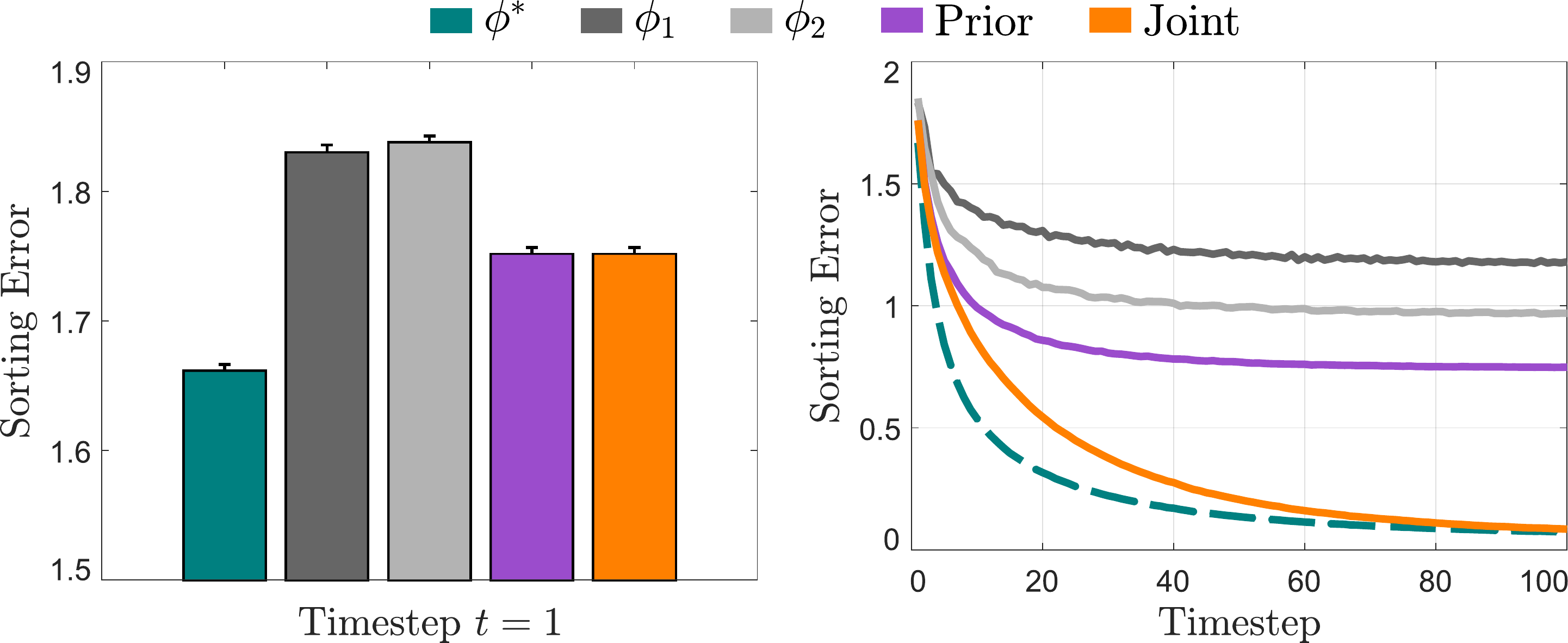}

		\caption{Learning when uncertain about the human's teaching strategy. Here two teaching strategies are equally likely. Left: number of screws incorrectly sorted after a single interaction. Right: number of screws incorrectly sorted after $t$ interactions. Maintaining a distribution over both teaching strategies results in more optimal robot behavior than always assuming one strategy. Notice that \emph{Prior} and \emph{Joint} are equal when $t=1$. Error bars indicate SEM.}

		\label{fig:learn1}
	\end{center}

\vspace{-0.5em}

\end{figure}

\begin{figure}[t]

	\begin{center}
		\includegraphics[width=1.0\columnwidth]{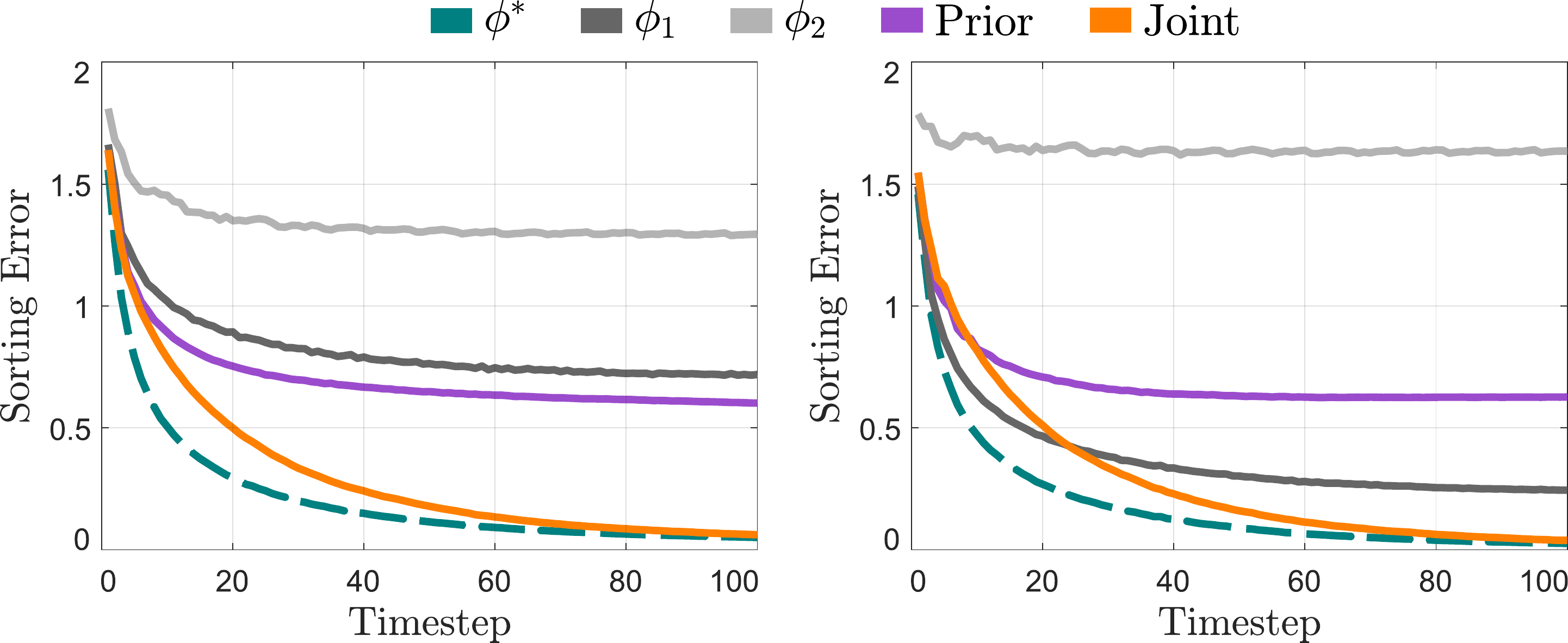}

		\caption{Learning when one human teaching strategy is more likely. Left: $70\%$ of humans use teaching strategy $\phi_1$. Right: $90\%$ of humans use $\phi_1$, but the robot incorrectly believes both strategies are equally likely \emph{a priori}, i.e., $b^0(\phi_1) = 0.5$. Even when the robot's prior is wrong, learning with strategy uncertainty (\emph{Joint}) leads to better performance over time than assuming $\phi_1$.}

		\label{fig:learn2}
	\end{center}

\vspace{-2em}

\end{figure}

\smallskip
\noindent \textbf{Simulation}. At timestep $t$ the robot observes the action $u^t$ and updates its belief $b^{t+1}(\theta)$ with (\ref{eq:L3}). Next, the robot optimally sorts $10$ screws based on its current belief \cite{ramachandran2007}. At timestep $t+1$ the task is repeated with the same end-user (who has a constant $\theta^*$ and $\phi^*$). The results of these simulations averaged across $10^5$ end-users are shown in Figs.~\ref{fig:learn1} and \ref{fig:learn2}.

\smallskip
\noindent \textbf{Analysis}. {Using our proposed \emph{Joint} observation model resulted in fewer errors than learning under $\phi_1$, $\phi_2$, or \emph{Prior}. With \emph{Joint} the robot was able to personalize its learning strategy to the current end-user across multiple iterations, and more accurately learn what the user was communicating (see Fig.~\ref{fig:learn1}). As expected, \emph{Prior} outperformed other fixed strategies when $b^0$ was correct; however, if the robot did not have an accurate prior over teaching strategies, the \emph{Prior} observation model (\ref{eq:L5}) was less optimal than $\phi_1$ (see Fig.~\ref{fig:learn2}, right). We found that our proposed approach was robust to this practical challenge: despite having the wrong prior, \emph{Joint} still caused the robot's behavior to converge to the ideal learner, $\phi^*$.}

\section{Robot Teaching with Strategy Uncertainty} \label{sec:teach}

Within this section we consider the opposite problem, where the expert robot is teaching the human about $\theta^*$, but does not know the end-user's learning strategy $\psi^*$. Teaching here is challenging because the robot is not certain what the user will learn from its actions. We first outline a specific instance of robot teaching, where the human learns through Bayesian inference. Next, we demonstrate how the robot can teach with multiple models of the human's learning strategy, and derive one solution to teaching with strategy uncertainty. In a simulated example, we compare these methods to robots that teach with a constant point estimate of $\psi^*$. We also describe how the robot can trade-off between teaching $\theta^*$ to and learning $\psi^*$ from the human via active teaching.

\subsection{Teaching Bayesian Humans} \label{sec:teachprob}

Similar to previous works in machine teaching \cite{huang2017,zhu2013} and cognitive science \cite{chater2008,jones2011}, we assume that the human learns by performing Bayesian updates. Thus, the human's belief over the target models after robot action $a^t$ becomes:
\begin{equation} \label{eq:T1}
	b^{t+1}(\theta) = \frac{b^t({\theta})\cdot \pi(a^t~|~x^t;\theta;\psi^*)}{Z(\psi^*)}
\end{equation}
where we use $;$ to denote that the human observes $\psi^*$ but not $\theta^*$. The denominator is again the normalizing constant:
\begin{equation} \label{eq:T2}
	Z(\psi) = \int_{\Theta} b^t(\xi) \cdot \pi(a^t ~ | ~ x^t ; \xi ; \psi)~d\xi
\end{equation}
We point out that $\pi$ in (\ref{eq:T1}) and (\ref{eq:T2}) is (\ref{eq:P2}), the policy that the human assigns to the robot. The human interprets the robot's actions---and updates its belief---based on this policy, which is parameterized by the human's true learning strategy $\psi^*$. Here $b^t$ is also the \emph{state} of the human at timestep $t$, and (\ref{eq:T1}) defines the \emph{state dynamics} (i.e., the human's transition function).

The robot should select actions so that this state transitions to $b(\theta^*)=1$. Let us define the ideal robot action as:
\begin{equation} \label{eq:T3}
	a^t = \text{arg}\max_{a} b^{t+1}(\theta^*)
\end{equation}
where $a^t$ will greedily maximize the human's belief in $\theta^*$ at the subsequent timestep. The human takes an action $u^t$ based on what they have previously learned; the human actions are therefore observations on the human's state, i.e., ${u^t = h(b^t)}$. For example, the human's action could be completing a test about the target models, or performing the task themselves. Here we consider the simplest case, where:
\begin{equation} \label{eq:T4}
	u^t = h(b^t) = b^t
\end{equation}
Hence, the human feedback $u^t$ provides their actual belief over the target models at the current timestep. The robot observes the human state $b^t$ from (\ref{eq:T4}), and selects action $a^t$ with (\ref{eq:T3}) to shift the human towards the desired state $b^{t+1}$.

\subsection{Multiple Learning Strategies}

Consider cases where the robot is teaching this Bayesian human, but does not know the human's learning strategy $\psi^*$. When $\psi^*$ (and therefore the future state $b^{t+1}$) is unknown, teaching is analogous to controlling an agent with unknown state dynamics \cite{astrom2008}. Define $\hat{b}$ as the robot's prediction of the human's state given the history of actions and world states:
\begin{equation} \label{eq:T5}
	\hat{b}^{t+1}(\theta) = P(\theta ~ | ~ u^{0:t},a^{0:t},x^{0:t})
\end{equation}
Since the human performs Bayesian inference (\ref{eq:T1}), and recalling that the robot observes $b^t(\theta)$, we equivalently have:
\begin{equation} \label{eq:T6}
	\hat{b}^{t+1}(\theta) = \int_{\Psi} \frac{u^t(\theta) \cdot \pi(a^t ~ | ~ x^t; \theta, \psi)}{Z(\psi)} \cdot b^t(\psi) ~ d\psi
\end{equation}
Within the above, $b(\psi)$ is the robot's belief over the human's learning strategies. For the state-of-the-art, the robot \emph{estimates} the human's learning strategy as $\psi^0$, so that (\ref{eq:T6}) reduces to:  
\begin{equation} \label{eq:T7}
	\hat{b}^{t+1}(\theta) = \frac{u^t(\theta) \cdot \pi(a^t ~ | ~ x^t; \theta, \psi^0)}{Z(\psi^0)}
\end{equation}
Instead, we here argue that the robot should teach with a \emph{belief over multiple learning strategies}. Let $b^0(\psi)$ be the prior over what $\psi^*$ is. If the robot never updates this initial belief, then the predicted human state after action $a^t$ becomes:
\begin{equation} \label{eq:T8}
	\hat{b}^{t+1}(\theta) = \int_{\Psi} \frac{u^t(\theta) \cdot \pi(a^t ~ | ~ x^t; \theta, \psi)}{Z(\psi)} \cdot b^0(\psi) ~ d\psi
\end{equation}
Comparing (\ref{eq:T8}) to (\ref{eq:T7}), now the robot reasons about how its actions are interpreted by each learning strategy. When selecting the action $a^t$ with (\ref{eq:T3})---where we replace $b^{t+1}$ with prediction $\hat{b}^{t+1}$---this robot teaches across multiple strategies.

\subsection{Inferring the Learning Strategy}

Because the robot is getting feedback from the user, however, we can also infer that specific user's learning strategy, $\psi^*$. Learning about $\psi^*$ provides a solution to teaching with strategy uncertainty (Definition 2), and results in robots that adapt their teaching to match the human. Let us formally define the robot's belief over learning strategies as:
\begin{equation} \label{eq:T9}
	b^{t}(\psi) = P(\psi ~ | ~ u^{0:t}, a^{0:t}, x^{0:t})
\end{equation}
We use the subscript $t$ instead of ${t+1}$ since $b^t(\psi)$ does not actually depend on $a^t$, as we will show. Applying Bayes' rule:
\begin{equation} \label{eq:T10}
	b^{t}(\psi) \propto P(u^{0:t} ~ | ~ a^{0:t}, x^{0:t}; \psi) \cdot P(\psi ~ | ~ a^{0:t}, x^{0:t})
\end{equation}
Recalling that the human's learning strategy is not altered by the robot, $P(\psi ~ | ~ a^{0:t}, x^{0:t}) = P(\psi)$. Moreover, because the human is a Bayesian learner, and $u^t = b^t$, here $u^t$ depends on $u^{t-1}$, $a^{t-1}$, $x^{t-1}$, and $\psi$ (\ref{eq:T1}). Hence, (\ref{eq:T10}) simplifies to:
\begin{equation} \label{eq:T11}
	b^{t}(\psi) \propto b^{t-1}(\psi) \cdot P\bigg[u^t ~ \bigg| ~ \frac{u^{t-1} \cdot \pi(a^{t-1} ~ | ~ x^{t-1}; \theta, \psi)}{Z(\psi)}\bigg]
\end{equation}
Intuitively, (\ref{eq:T11}) claims that the belief over learning strategies is updated based on the differences between the human's actual state (left side of the likelihood function) and the predicted human state given $\psi$ (right side of the likelihood function)\footnote{We used Kullback-Leibler (KL) divergence \cite{kullback1951} to define the likelihood of $u^t$ given the right side of (\ref{eq:T11}), but other options are possible.}. By observing $u$, we can use (\ref{eq:T11}) to infer the human's learning strategy. By then substituting (\ref{eq:T11}) back into (\ref{eq:T6}), the robot \emph{learns about $\psi^*$ while teaching the human $\theta^*\,$}: thus, using (\ref{eq:T6}) with (\ref{eq:T11}) addresses teaching with strategy uncertainty.

\subsection{Teaching Example}

\begin{figure}[t]

\vspace{0.5em}

	\begin{center}
		\includegraphics[width=0.7\columnwidth]{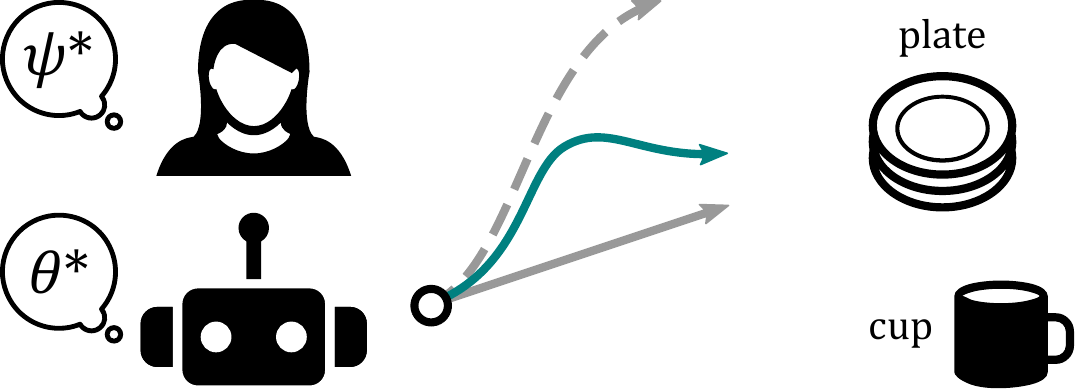}

		\caption{A robot is moving towards a goal position, and is trying to convey that goal to the human. Here the robot teacher is moving to the plate, $\theta^*$, but does not know the human's learning strategy, $\psi^*$. For example, the human may learn best from goal-directed trajectories (solid) or exaggerated trajectories (dashed). When the robot teacher reasons over each of these human learning strategies, it selects an action that teaches both types of learners (highlighted).}

		\label{fig:example2}
	\end{center}

\vspace{-2em}

\end{figure}

Here we provide an illustration of how reasoning over multiple learning models can improve teaching with uncertainty. As shown in Fig.~\ref{fig:example2}, the robot is moving towards goal position $\theta^*$, and wants to teach that goal to the nearby human. At each timestep $t$, the robot's action $a$ is an incomplete trajectory (e.g., see the three trajectory segments in Fig.~\ref{fig:example2}). After observing this robot trajectory, the human updates their belief over $\theta^*\,$; specifically, the human applies Bayesian inference to determine whether the robot's goal is the cup or the plate. The robot uses (\ref{eq:T3}) with prediction $\hat{b}^{t+1}$ to select the trajectory $a$ which will teach the human the most about $\theta^*$.

We consider two possible learning strategies $\psi \in \Psi$ for the simulated end-users. Humans with $\psi_1$ learn best from \emph{legible} (i.e., exaggerated) trajectories \cite{dragan2013}: $\pi(\psi_1) = (0.1, 0.3, 0.45)$ if the robot moves directly towards $\theta$, slightly exaggerates, or fully exaggerates, respectively. By contrast, under $\psi_2$ the user learns best from \emph{predictable} (i.e., goal-directed) trajectories, such that $\pi(\psi_2) = (0.35, 0.2, 0.15)$ if the robot moves directly towards $\theta$, slightly exaggerates, or exaggerates, respectively. The robot does not know which strategy a given user selects.

\smallskip
\noindent \textbf{Prediction Method}. We compare (\ref{eq:T6}), (\ref{eq:T7}), and (\ref{eq:T8}). Let $\psi_1$ denote a robot which predicts that every user learns with (\ref{eq:T7}), where $\psi^0 = \psi_1$. Likewise, $\psi_2$ is a robot that estimates $\psi^0 = \psi_2$. The \emph{Prior} robot reasons over both learning strategies using (\ref{eq:T8}), and our proposed \emph{Learn} robot solves teaching with strategy uncertainty by leveraging (\ref{eq:T6}) with (\ref{eq:T11}).

\begin{figure}[t]

\vspace{0.7em}

	\begin{center}
		\includegraphics[width=1.0\columnwidth]{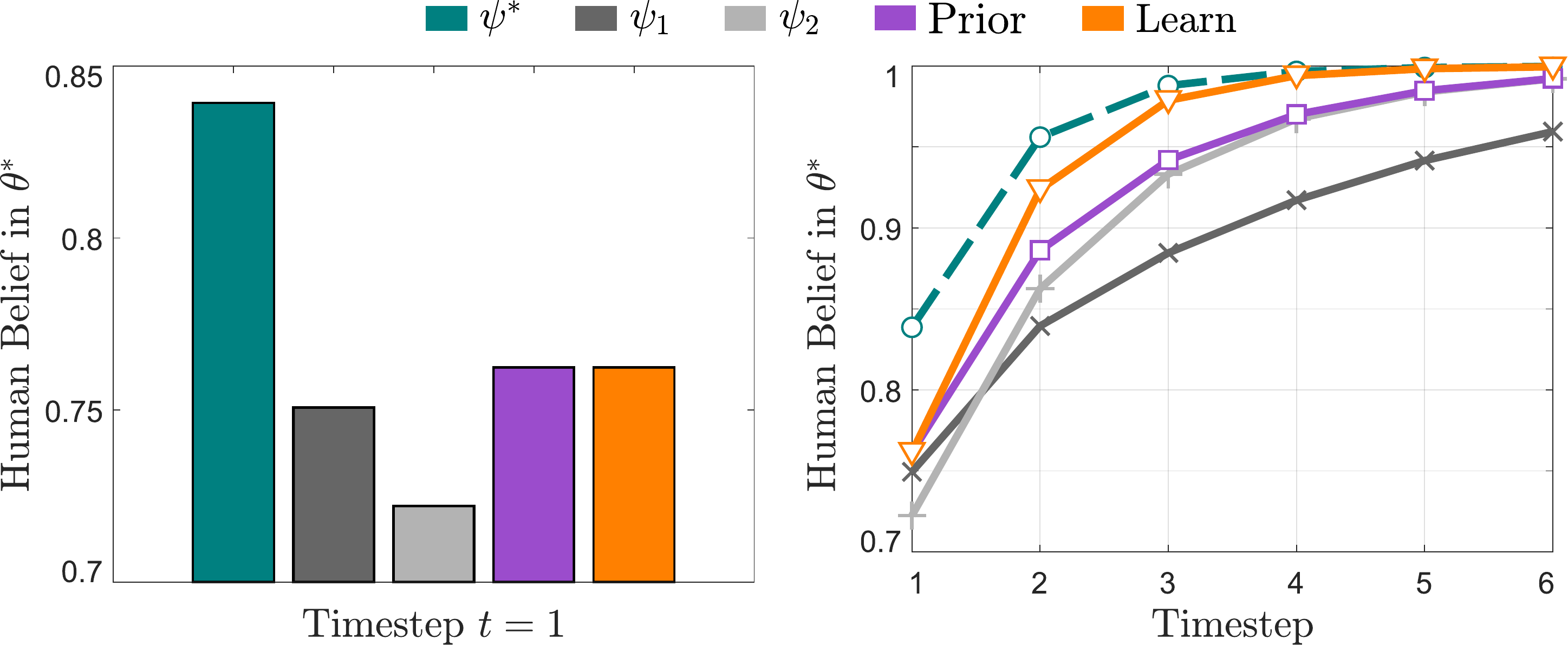}

		\caption{Teaching when uncertain about the human's learning strategy. Left: the human's confidence that the robot's goal is $\theta^*$ after one interaction. When using \emph{Prior} or \emph{Learn}, the robot selects a slightly exaggerated trajectory, which teaches both groups of learners. Right: the human's belief after $t$ iterations. When the robot reasons about a distribution over learning strategies, it teaches $\theta^*$ more quickly than when focusing on a single type of learner ($\psi_1$ and $\psi_2$).}

		\label{fig:teach1}
	\end{center}

\vspace{-0.5em}

\end{figure}

\begin{figure}[t]

	\begin{center}
		\includegraphics[width=1.0\columnwidth]{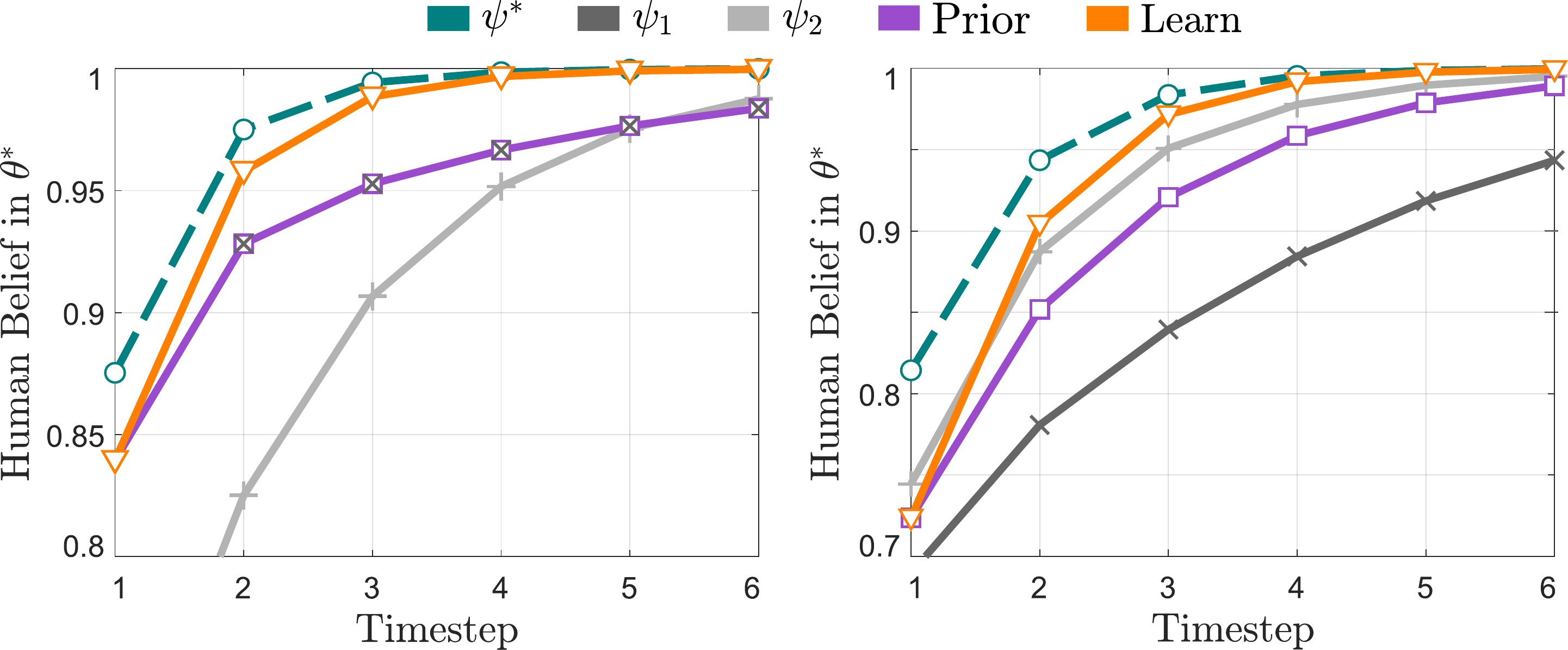}

		\caption{Teaching when one learning strategy is more likely. Left: the robot knows that $80\%$ of users learn with $\psi_1$, and so the \emph{Prior} robot greedily selects legible actions (\emph{Prior}$=\psi_1$). Right: the robot initially thinks both strategies are equally likely, but $70\%$ of users have $\psi_2$. Learning about $\psi^*$ while teaching $\theta^*$ (\emph{Learn}) outperforms $\psi_2$ over time, even with the wrong prior.}

		\label{fig:teach2}
	\end{center}

\vspace{-2em}

\end{figure}

\smallskip
\noindent \textbf{Simulation}. The robot observes the human action $u$---i.e., the human's current belief---and selects an action $a$ using (\ref{eq:T3}) and its prediction method. The robot can select between $6$ different legible or goal-directed trajectory segments ($3$ for each goal $\theta$). The human is a Bayesian learner. Our results (averaged across $10^5$  simulated users) are depicted in Figs.~\ref{fig:teach1} and \ref{fig:teach2}.

\smallskip
\noindent \textbf{Analysis}. {Robots using our proposed \emph{Learn} approach more quickly taught $\theta^*$ than with the fixed teaching methods $\psi_1$, $\psi_2$, or \emph{Prior}. Reasoning over human learning strategies led to better teaching during a single interaction (see Fig.~\ref{fig:teach1}). For multiple iterations, we tested practical scenarios where the robot has the wrong prior: in every case, \emph{Learn} yielded the fastest convergence, and taught as well as the ideal teacher after $\approx 5$ timesteps (see Fig.~\ref{fig:teach2}). Intuitively, the \emph{Learn} robot gradually shifted to teaching with either legible or predictable trajectories, while the \emph{Prior} robot continued to compromise between both strategies instead of adapting to the specific user.}

\subsection{Active Teaching}

Like we saw in the previous example, learning about the human's learning strategy $\psi^*$ can improve the robot's teaching. Hence, we here focus on selecting robot actions which \emph{actively} gather information about $\psi^*$, so that the robot more quickly adapts its teaching to the end-user. Let us formulate teaching with strategy uncertainty as a partially observable Markov decision process (POMDP) \cite{russell2009}: the state is $\big(b^t({\theta}), \theta^*, \psi^*\big)$, the action is $(a^t,x^t)$, the observation is $u^t$, the state transitions with (\ref{eq:T1})---where $\theta^*$ and $\psi^*$ are constant---the observation model is (\ref{eq:T11}), and the reward is $b^t(\theta^*)$. Solving this POMDP causes the robot to optimally trade-off between exploring for more information about $\psi^*$ and exploiting that information to maximize the human's belief in $\theta^*$. When solving this POMDP is intractable, we can more simply perform active teaching by favoring actions that gather information about $\psi^*$ \cite{sadigh2016iros}:
\begin{equation} \label{eq:T12}
	a^t = \text{arg}\max_{a} \big\{b^{t+1}(\theta^*) - \lambda \cdot H(b^{t+1}(\psi))\big\}
\end{equation}
In the above, $\lambda \geq 0$, and $H$ is the Shannon entropy. Comparing (\ref{eq:T12}) to (\ref{eq:T3}), now the robot selects actions to disambiguate between the possible learning strategies (i.e., reduce the entropy of the robot's belief over $\psi$). Intuitively, we expect a robot that is actively teaching with (\ref{eq:T12}) to select actions, $a$, which \emph{cause users with different learning strategies to respond in different ways}, allowing that robot to more easily infer $\psi^*$.

\section{Robot Learning Simulations} \label{sec:experiment}

To compare our \emph{learning with strategy uncertainty} against the state-of-the-art in a realistic problem setting, we performed a simulated user study. We here consider an instance of inverse reinforcement learning (IRL): the human demonstrates a policy, and the robot attempts to infer the human's reward function from that demonstrated policy \cite{ng2000,abbeel2004,osa2018}. Unlike the example in Section~\ref{sec:learnex}, now $\theta^*$ (the human's reward parameters) and $\phi^*$ (the human's demonstration strategy) lie in continuous spaces. We compared robots that learn $\theta^*$ with a constant point estimate of $\phi^*$ to our proposed method, where the robot learns about both $\theta^*$ and $\phi^*$ from the human. {To test the robustness of our method within more complex and challenging scenarios, we also introduced noisy end-users, who did not follow any of the modeled teaching strategies.}

\subsection{Setup and Simulated Users}

Within each simulation the human and robot were given an 8-by-8 gridworld (64 states). The state reward, $R(x,\theta)$, is the linear combination of state features $f(x)$ weighted by $\theta$, so that $R(x,\theta) = \theta \cdot f(x)$. The human knows $\theta^*$, and provides a demonstration $\pi(u ~ | ~ x, \theta^*, \phi^*)$. This demonstration is a policy, where the human labels each state $x$ with action $u$; actions deterministically move in one of the four cardinal directions. The discount factor---which defines the relative importance of future and current rewards---was fixed at $\gamma = 0.9$.

\begin{figure}[t]

\vspace{0.6em}

	\begin{center}
		\includegraphics[width=0.95\columnwidth]{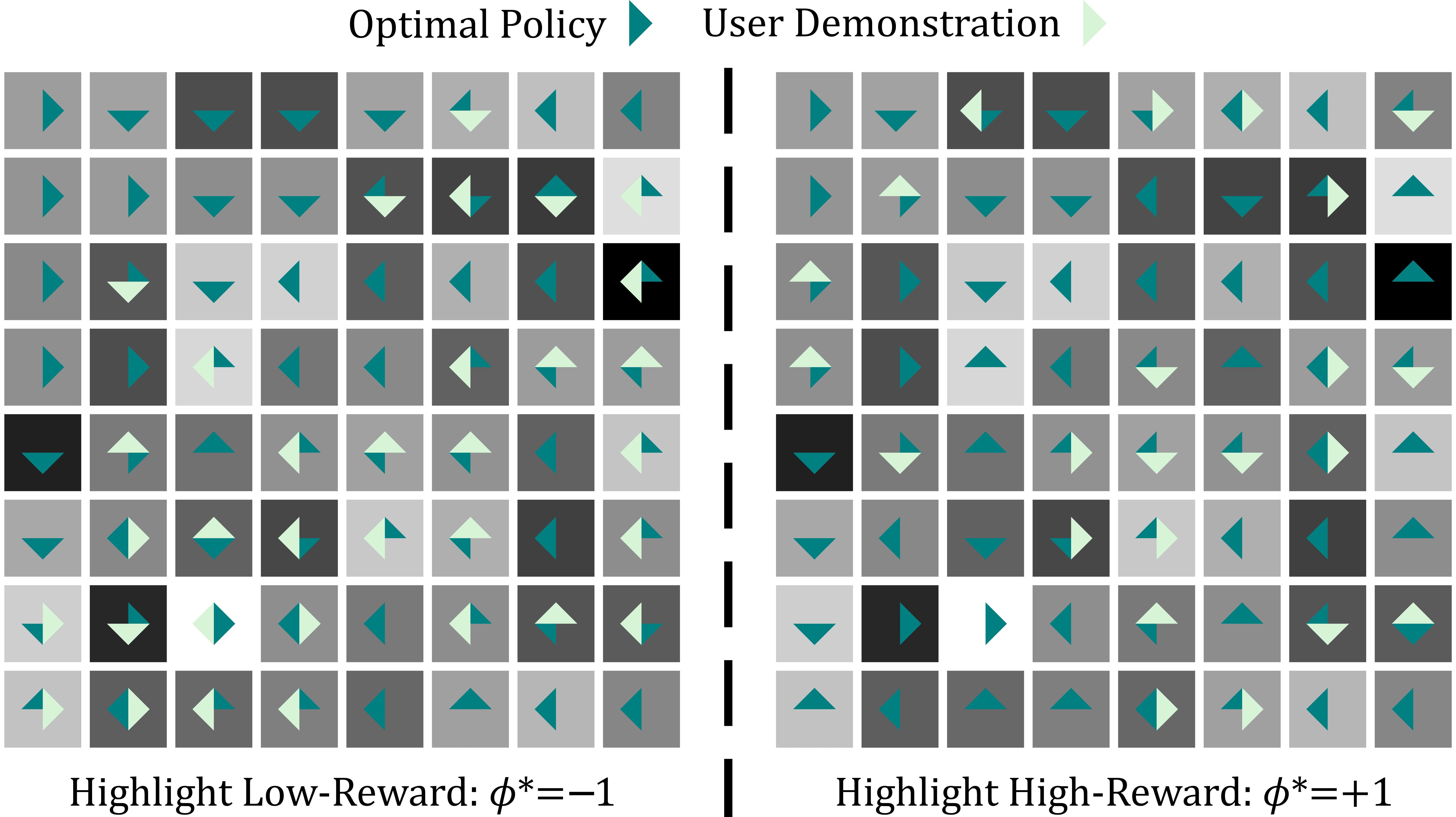}

		\caption{Sample 8-by-8 gridworld labeled by two different simulated users. Both users have the same reward parameters $\theta^*$, and the grid cells are colored based on this reward (lighter cells have higher reward). The optimal policy for $\theta^*$ is shown by the darker arrows. Left: the human leverages teaching strategy $\phi^* = -1$, and teaches the robot by biasing their demonstration to highlight low-reward states. Right: this user instead uses teaching strategy $\phi^* = +1$, and favors states with locally higher reward. The robot attempts to learn $\theta^*$ given a demonstration (like the ones shown above), but does not know $\phi^*$.} 

		\label{fig:mdp}
	\end{center}

\vspace{-2em}

\end{figure}

Our setting is based upon previous IRL works \cite{osa2018}, where this problem is more formally introduced as a Markov decision process (MDP). These prior works typically assume that the human's demonstrated policy approximately solves the MDP, i.e., maximizes the expected sum of discounted rewards \cite{ramachandran2007,ziebart2008}. By contrast, \emph{we here considered users with a spectrum of demonstration strategies}. Let $Q(x,u,\theta)$ be the reward for taking action $u$ in state $x$, and then following the optimal policy for reward parameters $\theta$. We define the probability that the simulated user takes action $u$ given $x$, $\theta^*$, and $\phi^*$ as:
\begin{equation} \label{eq:S1}
	\pi \propto \exp\Big\{\alpha \Big[Q(x,u,\theta^*) + \phi^* \big(R(x',\theta^*) - R(x,\theta^*)\big)\Big]\Big\}
\end{equation}
where $\phi^* \in [-1,1]$, and $x'$ is the state reached after taking action $u$ in state $x$. When $\phi^* = 0$, (\ref{eq:S1}) is the same as the observation model from \cite{ramachandran2007,ziebart2008}. As $\phi^* \rightarrow +1$, the human biases their demonstration towards states that have locally higher rewards; conversely, when $\phi^* \rightarrow -1$, the human favors states with lower rewards. Sample user demonstrations with different teaching strategies are shown in Fig.~\ref{fig:mdp}.

\subsection{Independent Variables}

We compared four different approaches for learning $\theta^*$ from the user's demonstration: $\phi^*$, $\phi = -1$, $\phi = +1$, and \emph{Joint}. Under $\phi^*$ the ideal robot knows the human's true teaching strategy, while $\phi = -1$ and $\phi = +1$ indicate robots which assume that the human's demonstration is biased towards low-reward or high-reward states, respectively. \emph{Joint} refers to a robot which attempts to learn both $\phi^*$ and $\theta^*$ from the human's demonstration, as discussed in Section~\ref{sec:joint}.

To see how these approaches scale with the length of the feature vector, $f \in F$, we performed simulations with $|F| = 4$, $8$, and $16$ features. In practice, each state $x$ was randomly assigned a feature vector with $|F|$ binary values, indicating which features were present in that particular gridworld state.

Finally, to test how well the robot learned when the human demonstrations were imperfect, we varied the value of $\alpha$ in (\ref{eq:S1}). Parameter $\alpha$ represents how close to optimal the human is: as $\alpha \rightarrow 0$, the human becomes increasingly random, while the human always chooses the best action when $\alpha \rightarrow \infty$.

We simulated $100$ users for each combination of $|F|$ and $\alpha$, where the users' teaching strategies were uniformly distributed in the continuous interval $\phi^* \in [-1,1]$. The gridworld and $\theta^*$ were randomly generated for each individual user.

\subsection{Dependent Measures}

For each simulation we measured the robot's learning performance in terms of \emph{Reward Error}, \emph{Strategy Error}, and \emph{Policy Loss}. \emph{Reward Error} is the difference between the robot's mean estimate of $\theta^*$ and the correct reward parameters: $\| \theta^* - \hat{\theta} \|_1$. Similarly, \emph{Strategy Error} is the error between the robot's mean estimate of $\phi^*$ and the user's actual teaching strategy: $| \phi^* - \hat{\phi}|$. \emph{Policy Loss} measures how much reward is lost by following the robot's learned policy (which maximizes reward under $\hat{\theta}\,$) as compared to the optimal policy for $\theta^*$ \cite{ramachandran2007}. The code for our examples and simulations can be found at \url{https://github.com/dylanplosey/iact_strategy_learning}.



\subsection{Results and Discussion}

\begin{figure*}[t]

\vspace{0.6em}

	\begin{center}
		\includegraphics[width=0.835\textwidth]{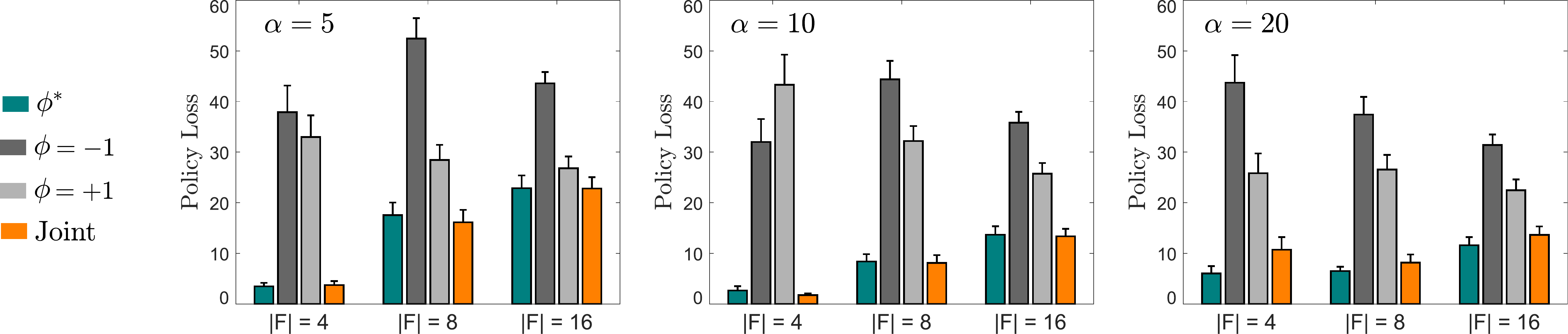}

		\caption{Learning the human's teaching strategy in addition to learning the human's reward parameters (\emph{Joint}) reduces the robot's \emph{Policy Loss}. Left: $\alpha = 5$. Middle: $\alpha = 10$. Right: $\alpha = 20$. Recall that $\phi^*$ is an ideal learner, and lower values of $\alpha$ indicate increasingly random users. Error bars denote SEM.}

		\label{fig:policy}
	\end{center}

\vspace{-0.7em}

\end{figure*}

\begin{figure*}[t]

	\begin{center}
		\includegraphics[width=0.835\textwidth]{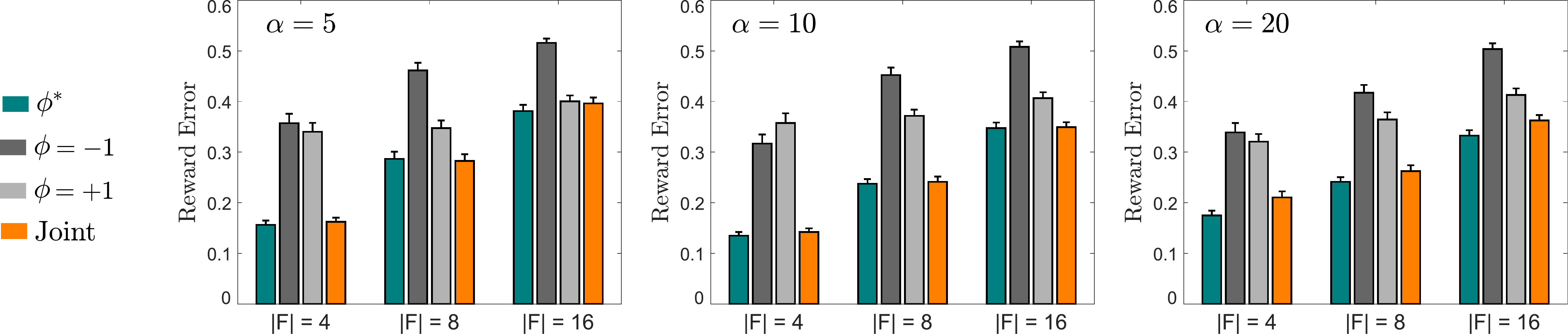}

		\caption{The robot more accurately learns the human's reward parameter $\theta^*$ when maintaining a distribution over the teaching strategies. Left: $\alpha = 5$. Middle: $\alpha = 10$. Right: $\alpha = 20$. Note that $\phi = +1$ outperformed $\phi = -1$ because it could correctly interpret the demonstrations from a wider range of $\phi^*$.}

		\label{fig:reward}
	\end{center}

\vspace{-0.7em}

\end{figure*}

We performed a mixed ANOVA with the number of features and value of $\alpha$ as between-subjects factors, and the learning approach as a within-subjects factor, for both \emph{Policy Loss} and \emph{Reward Error} (see Figs.~\ref{fig:policy} and \ref{fig:reward}). Since we found a statistically significant interaction for both dependent measures ($p<.05$), we next determined the simple main effects.

Simple main effects analysis showed that \emph{Joint} resulted in significantly less \emph{Policy Loss} than either $\phi = -1$ or $\phi = +1$ for each different combination of $|F|$ and $\alpha$ ($p<.05$). We similarly found that \emph{Joint} resulted in significantly less \emph{Reward Error} ($p<.001$) for every case except $|F| = 16$, $\alpha = 5$; here there was no statistically significant difference between \emph{Joint} and $\phi = +1$ (${p = .498}$). These results from Figs.~\ref{fig:policy} and \ref{fig:reward} suggest that learning while maintaining a distribution over $\phi$ results in objectively better performance than learning with a fixed point estimate of $\phi^*$.

\begin{figure}[t]

	\begin{center}
		\includegraphics[width=1.0\columnwidth]{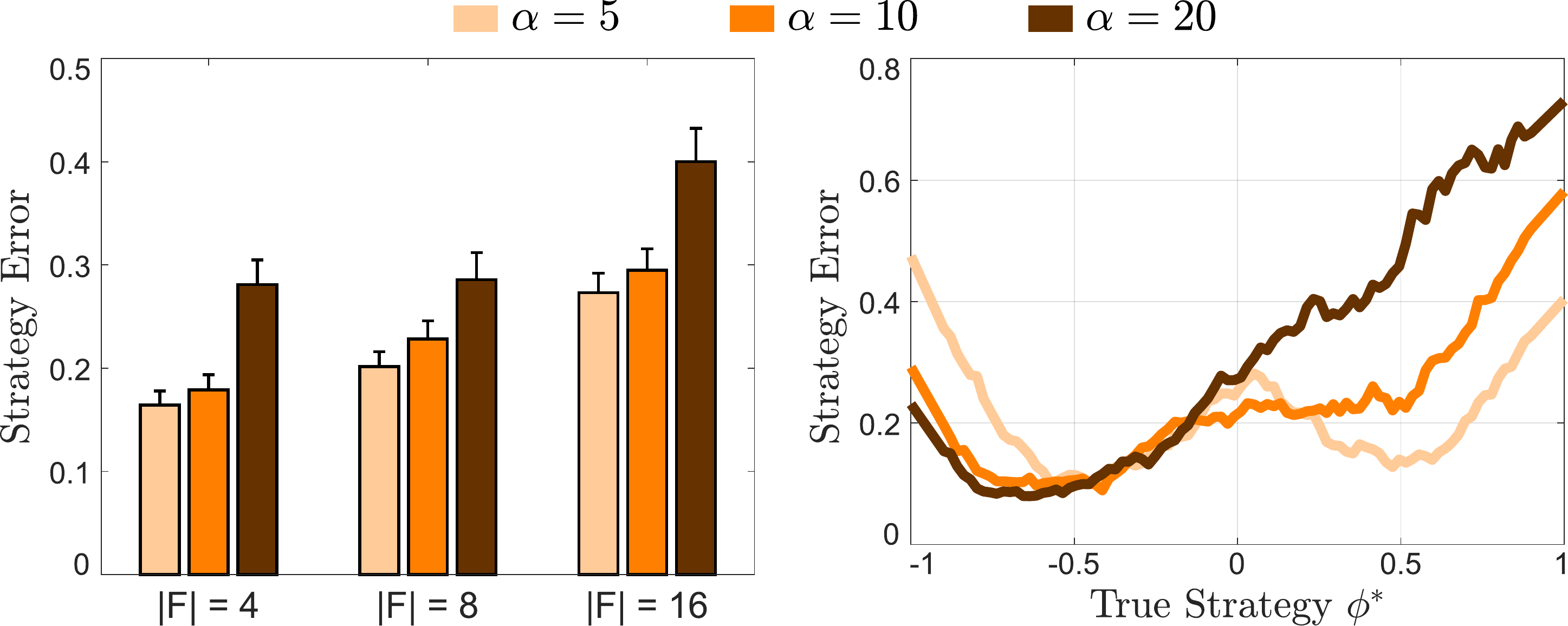}

		\caption{Error between the learned and true teaching strategies  under \emph{Joint}. Left: \emph{Strategy Error} for different numbers of features and values of $\alpha$. Right: \emph{Strategy Error} (averaged across features) as a function of the user's teaching strategy. Teachers with $\phi^* < 0$ were easier to distinguish than with $\phi^* > 0$.}
        
		\label{fig:phi}
	\end{center}

\vspace{-0.5em}

\end{figure}

Next, we investigated how well the \emph{Joint} method learned the individual users' teaching strategies. We performed a two-way ANOVA to find the effects of $|F|$ and $\alpha$ on the \emph{Joint} robot's \emph{Strategy Error} (see Fig.~\ref{fig:phi}). We found that the number of features ($F(2,891)=23.813,p<.001$) and the human's $\alpha$ ($F(2,891)=22.679,p<.001$) had a significant main effect. Post-hoc analysis with Tukey HSD revealed that $|F| = 16$ and $\alpha = 20$ led to significantly higher \emph{Strategy Error} than the other values of $|F|$ and $\alpha$, respectively. As shown in Fig.~\ref{fig:phi}, the robot had larger \emph{Strategy Error} for higher values of $\alpha$ because it was unable to distinguish between teachers with $\phi^* > 0$; i.e., these different teachers provided similar policy demonstrations when $\alpha = 20$.

\begin{figure}[t]

	\begin{center}
		\includegraphics[width=1.0\columnwidth]{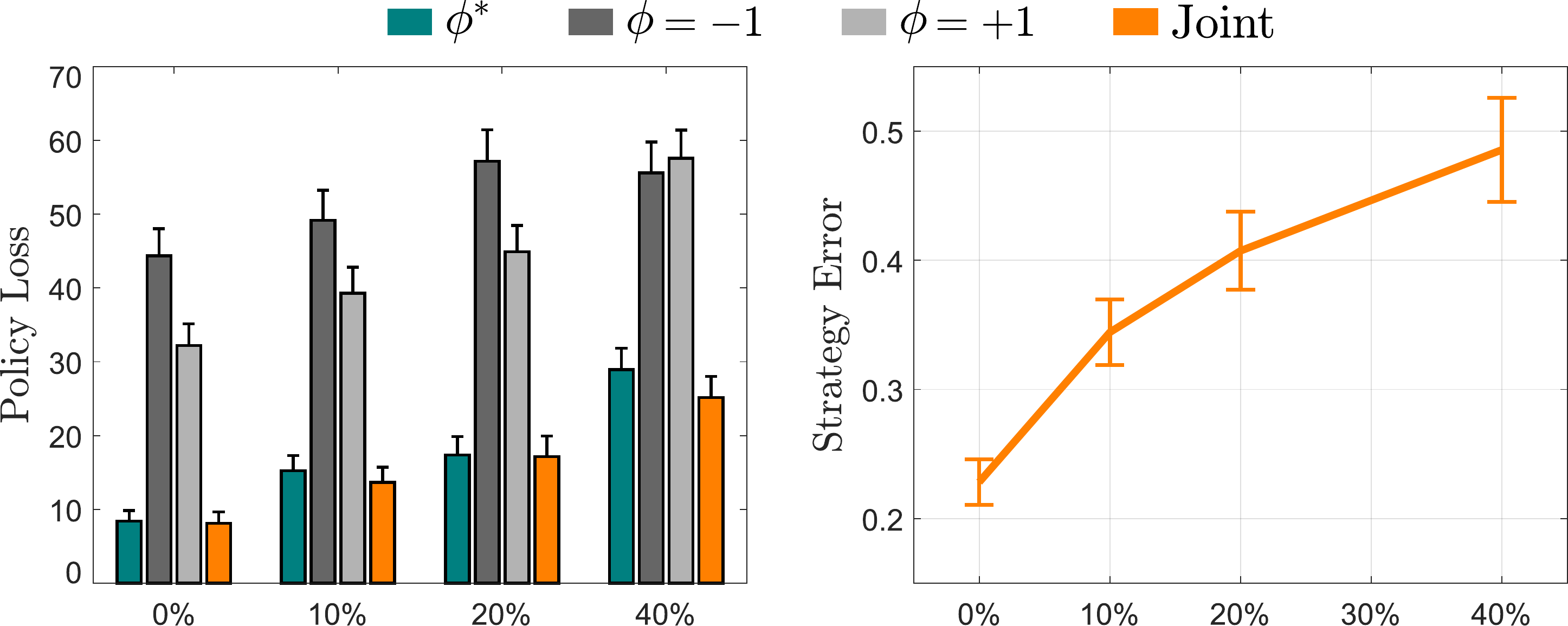}

		\caption{Learning from users who provided demonstrations with unmodeled noise. Left: \emph{Policy Loss} for each learning approach as the noise ratio increased. Right: \emph{Strategy Error} (for \emph{Joint}) as a function of the noise ratio. Although the \emph{Joint} robot's \emph{Strategy Error} increased in proportion to the noise ratio during the user demonstrations, its \emph{Policy Loss} was on par with the ideal learner, $\phi^*$.}
        
		\label{fig:noise}
	\end{center}

\vspace{-2em}

\end{figure}

Finally, we conducted a followup simulation in which we introduced unmodeled noise (see Fig.~\ref{fig:noise}). Here $|F|=8$ and $\alpha = 10$, but we now increased the ratio of the human taking completely random actions, which were not modeled in (\ref{eq:S1}). \emph{Joint} resulted in significantly less \emph{Policy Loss} than $\phi = -1$ or $\phi = +1$, even as the ratio of unmodeled user noise increased. Hence, reasoning over multiple strategies still improved performance for cases where the noisy end-user did not comply with any of the modeled teaching strategies.

\subsection{{Challenges and Limitations}}

{Although this simulated user-study supports learning with strategy uncertainty, there are still practical challenges that may limit our proposed approach. In particular, the end-user's actual interactions may not match any of the learning or teaching models, such that $\phi^* \notin \Phi$ or $\psi^* \notin \Psi$. Having the wrong hypothesis space is often unavoidable when using models to learn from humans: but, as we show in Fig.~\ref{fig:noise}, our proposed approach does remain robust to some errors in the hypothesis space (such as noisy users, that do not follow any of the included models). In practice, designers could leverage data from previous trials to construct a richer space of possible interaction strategies, so that $\Phi$ is updated to include $\phi^*$.}

\section{Conclusion} \label{sec:conclusion}

Because the human's interaction strategy during HRI varies from end-user to end-user, robots that assume a fixed, pre-defined interaction strategy may result in inefficient, confusing interactions. Thus, we proposed that the robot should maintain a distribution over the human interaction strategies, and \emph{exchange information while reasoning over this distribution}. We here introduced robot (a) learning with strategy uncertainty and (b) teaching with strategy uncertainty, and derived solutions to both novel problems. We performed learning and teaching examples---as well as learning simulations---and compared our approach to the state-of-the-art. Unlike standard approaches that assume every user interacts in the same way, we found that attempting to infer each individual end-user's interaction strategy led to improved robot learning and teaching, {while remaining robust to unmodeled strategies.}


\bibliographystyle{IEEEtran}
\bibliography{IEEEabrv,BibFile}

\end{document}